\documentclass[conference]{IEEEtran}
\IEEEoverridecommandlockouts

\usepackage{cite}
\usepackage{amsmath,amssymb,amsfonts}
\usepackage{algorithmic}
\usepackage{graphicx}
\usepackage{textcomp}
\usepackage{xcolor}
\usepackage{subcaption}%
\usepackage{booktabs}%
\usepackage[T1]{fontenc}       
\usepackage{newtxtext,newtxmath}

\def\BibTeX{{\rm B\kern-.05em{\sc i\kern-.025em b}\kern-.08em
    T\kern-.1667em\lower.7ex\hbox{E}\kern-.125emX}}
\begin{document}


\title{TextMamba: Scene Text Detector with Mamba}

\author{\IEEEauthorblockN{1\textsuperscript{st} Qiyan Zhao}
\IEEEauthorblockA{\textit{Fujian Key Laboratory of Pattern} \\ 
\textit{Recognition and Image Understanding,} \\
\textit{Xiamen University of Technology}\\
Xiamen, China \\
qiyanzhao618@gmail.com}
\and
\IEEEauthorblockN{2\textsuperscript{nd} Yue Yan}
\IEEEauthorblockA{\textit{Department of Automatic Control}\\
\textit{and Systems Engineering,} \\
\textit{The University of Sheffield} \\
Sheffield, England \\
y20000525@yeah.net}
\and
\IEEEauthorblockN{3\textsuperscript{rd} Da-Han Wang}
\IEEEauthorblockA{\textit{Fujian Key Laboratory of Pattern} \\ 
\textit{Recognition and Image Understanding,} \\
\textit{Xiamen University of Technology}\\
Xiamen, China \\
wangdh@xmut.edu.cn}
}

\maketitle

\begin{abstract}

In scene text detection, Transformer-based methods have addressed the global feature extraction limitations inherent in traditional convolution neural network-based methods. However, most directly rely on native Transformer attention layers as encoders without evaluating their cross-domain limitations and inherent shortcomings: forgetting important information or focusing on irrelevant representations when modeling long-range dependencies for text detection. The recently proposed state space model Mamba has demonstrated better long-range dependencies modeling through a linear complexity selection mechanism. Therefore, we propose a novel scene text detector based on Mamba that integrates the selection mechanism with attention layers, enhancing the encoder's ability to extract relevant information from long sequences. We adopt the Top\_k algorithm to explicitly select key information and reduce the interference of irrelevant information in Mamba modeling. Additionally, we design a dual-scale feed-forward network and an embedding pyramid enhancement module to facilitate high-dimensional hidden state interactions and multi-scale feature fusion. Our method achieves state-of-the-art or competitive performance on various benchmarks, with F-measures of 89.7\%, 89.2\%, and 78.5\% on CTW1500, TotalText, and ICDAR19ArT, respectively. Codes will be available.

\end{abstract}


\section{Introduction}
Scene text detection is a critical task in computer vision, given its abundant application scenarios, such as office automation and autonomous driving. Recent approaches primarily follow either a segmentation-based or regression-based pipeline. In segmentation-based methods, pixel-level predictions are grouped using elaborate post-processing algorithms. For example, PSENet \cite{PSE} proposes a progressive scale expansion algorithm to aggregate text instances pixels. PAN \cite{1} and PAN++ \cite{2} propose a learnable post-processing based on PSENet \cite{PSE}. DB \cite{DB} and DB++ \cite{3} propose a differentiable binarization module to optimize post-processing. Alternatively, regression-based methods model bounding boxes using heuristic representations. ABCNet \cite{5} and ABCNetV2 \cite{4} employ parameterized Bezier curves to fit arbitrarily shaped text boundaries. CRAFT \cite{22} combines character-level predictions into connected regions to represent overall text instances. TextSnake \cite{6} represents textual geometric properties through a sequence of ordered disks.

\begin{figure}[htb]
\centering
\centerline{\includegraphics[width=8.5cm]{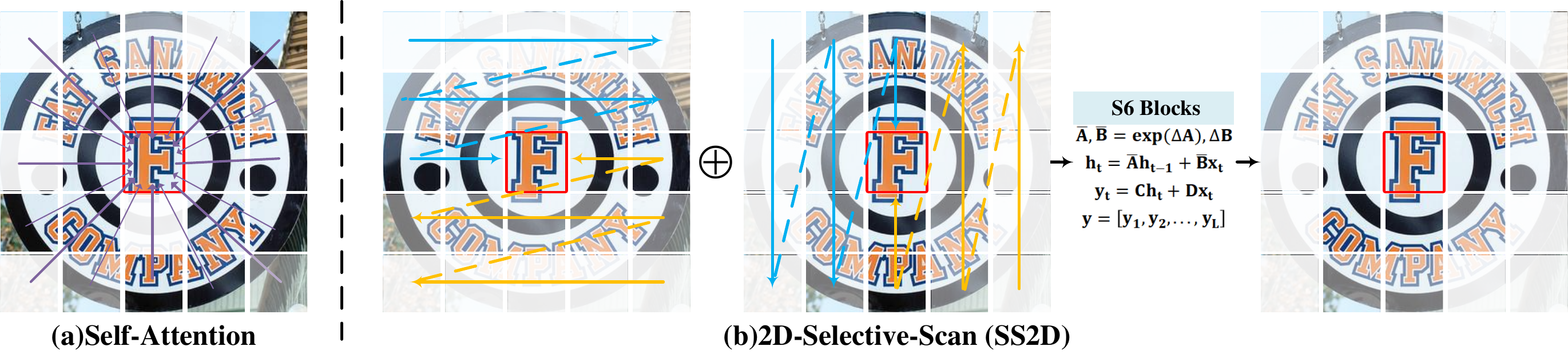}}
\caption{Comparison of Self-Attention \cite{8} and Mamba's SS2D \cite{15}. The red box indicates the query patch, and the patch transparency indicates the corresponding of degree information loss. $x, y$ in S6 block are input and output variables respectively and $\bar{A}$, $\bar{B}$, $\bar{C}$ are learnable parameters.}
\label{fig:res}
%
\end{figure}

Most of the aforementioned methods \cite{PSE}-\cite{6} rely on the local feature extraction capabilities of convolution neural network (CNN), which inherently limits their ability to capture global features in images. To address this limitation, recent studies have incorporated Transformer architectures with their core self-attention mechanisms \cite{att}-\cite{10} into the scene text detection domain. For example, \cite{8},\cite{9} introduced the Transformer's attention layers into the scene text detection framework to model global contextual features. Building on Detection Transformer (DETR) family models \cite{tr}-\cite{10}, some studies \cite{testr}-\cite{12} exploit different end-to-end 
Transformer-based text detection methods.

Nevertheless, most previous methods \cite{8}-\cite{12} directly use native self-attention in Transformers as encoders without fully assessing its cross-domain effectiveness. Specifically, challenges arise from the similarity of text instances to their surroundings and their arbitrary aspect ratios, which complicates the extraction of relevant and crucial information from long sequences. Standard self-attention often lacks concentration when modeling global feature dependencies, leading to forgetting important information or focusing on irrelevant representations, particularly in longer sequences \cite{et}-\cite{14}.

Recently proposed state space models (SSMs), such as Mamba models \cite{vm}-\cite{15}, effectively model long-range dependencies with linear complexity. The core 2D selective scan model (SS2D) within Mamba \cite{15} excels in remembering relevant information over long sequences via its implicit selection mechanism. Consequently, we attempt to leverage Mamba's advancements to address the limitations of the native Transformer encoder and enhance text detection performance.

In this paper, we propose a novel scene text detector with Mamba, integrating the SS2D within the encoder to capitalize on Mamba's strengths. Specifically, the encoder initially employs deformable attention layers \cite{10} to capture global features, applying the Top\_k algorithm to explicitly sparsify attention weights and eliminate distractions from irrelevant information. Subsequently, we incorporate the SS2D into the encoder, constructing a hybrid encoder mixed with Mamba and attention layers. To enhance the interaction of high-dimensional hidden state information within the encoder, we propose a dual-scale feed-forward network (DSFFN) to replace the traditional feed-forward network (FFN). This composite encoder unit, integrating these three designs, is termed a Mix-SSM block. Additionally, we design an embedding pyramid enhancement module (EPEM) to optimize multi-scale feature fusion. In summary, the primary contributions of our method are as follows:

\begin{itemize}
\item 
This paper proposes a novel scene text detector, named TextMamba, which models long-range dependencies more effectively. We analyze the limitations of Transformer and introduce Mamba's selection mechanism to improve the performance of model scene text detection.

\item 
TextMamba introduces the Mix-SSM blocks on the basic structure of the DETR. The encoder part uses the deformable attention layer and explicitly selects key information by the Tok\_k algorithm. The obtained token sequences are passed into SS2D to further model the long-range dependencies. Mix-SSM also employs the DSFFN module to implicitly enhance information interactions. Additionally, we also propose an EPEM module for token-wise multi-scale feature fusion.

\item 
We perform extensive experiments and ablations analysis in several benchmarks. Our experiments show consistent and excellent performance under three different benchmarks. The F-measure is 89.7\% on CTW150, 89.2\% on TotalText, and 78.5\% on ICDAR19 ArT, respectively.
\end{itemize}

\section{Related Work}

\subsection{Scene Text Detection}

\subsubsection{CNN-based Method}

Due to their exceptional ability to extract semantic information and local features, convolutional neural networks (CNNs) have become the foundational model architecture for computer vision tasks. Most prior scene text detection methods \cite{PSE}-\cite{6} leverage CNNs to extract feature maps from images, followed by various pipelines to predict text regions. For instance, by integrating an object detection framework, CTPN \cite{ctpn} and TextBoxes \cite{Textboxes} decode the feature maps into coordinates of horizontal text instances. EAST \cite{east} introduces an end-to-end pixel-level regression method based on CNNs to better accommodate multi-oriented text. To better handle arbitrarily shaped text instances, later studies \cite{4,5,22,6} proposed a series of heuristic representations for processing regression bounding boxes.

Several studies \cite{1,2,3,DB,PSE} have utilized segmentation heads for pixel-level predictions from feature maps extracted by CNNs, employing sophisticated post-processing methods to refine predicted regions. For example, PSE \cite{PSE} and PAN \cite{2,3} predict text kernels and apply adaptive scaling algorithms. DB \cite{3} and DB++ \cite{3} introduce deformable convolutional networks \cite{dc} to enhance the prediction of arbitrarily shaped text instances. However, due to the limited global feature extraction capability of CNNs, both segmentation-based and regression-based methods often exhibit suboptimal performance.

\subsubsection{Transformer-based Method}
To overcome the limitations of CNNs in global feature extraction, several studies have introduced the Transformer \cite{7, 10} framework to model global information. DTTR \cite{9} utilizes the Transformer's decoder to capture global context information from feature maps extracted by CNNs. DEtection TRansformer (DETR) \cite{tr} establishes a fully end-to-end Transformer-based object detection framework, simplifying complex post-processing steps. Building on DETR, TESTR \cite{testr} employs an encoder-decoder structure to predict bounding boxes. Similarly, DPText \cite{11} transforms CNN-extracted feature maps into learnable queries that are processed by the DETR framework, predicting text instance regions through a refined decoding step. While these methods leverage the attention mechanism to capture global features and semantic dependencies, few studies have thoroughly examined the limitations of the attention layer in the context of scene text detection.

\subsection{State Space models}\label{AA}

State Space Models (SSMs), such as the S4 model \cite{13} and Mamba \cite{14}, have recently demonstrated promising capabilities in modeling long-range dependencies. Notably, compared to the attention mechanism of Transformers, the selection mechanism of the Mamba model offers significant advantages in modeling long-range dependencies across various natural language processing tasks. The integration of the SS2D module in VMamba \cite{15} has successfully incorporated Mamba into vision backbones. Subsequent studies \cite{15}-\cite{16} have validated the effectiveness of SS2D across a range of computer vision tasks, including object detection \cite{15}, medical image processing \cite{vm2}, image reconstruction \cite{vm3}, and semantic segmentation \cite{16}.

As depicted in Fig. 1, unlike traditional self-attention that computes attention scores for each query patch relative to the entire image, SS2D utilizes a 2D scanning selection mechanism that sequentially traverses patches along four predefined paths, allowing for the indefinite retention of relevant information. The scan sequence $x$ is input into the S6 Block for feature extraction, with the resulting output sequences $y$ subsequently merged into the output map.

\begin{figure*}[h]
\centering
\centerline{\includegraphics[width=16cm]{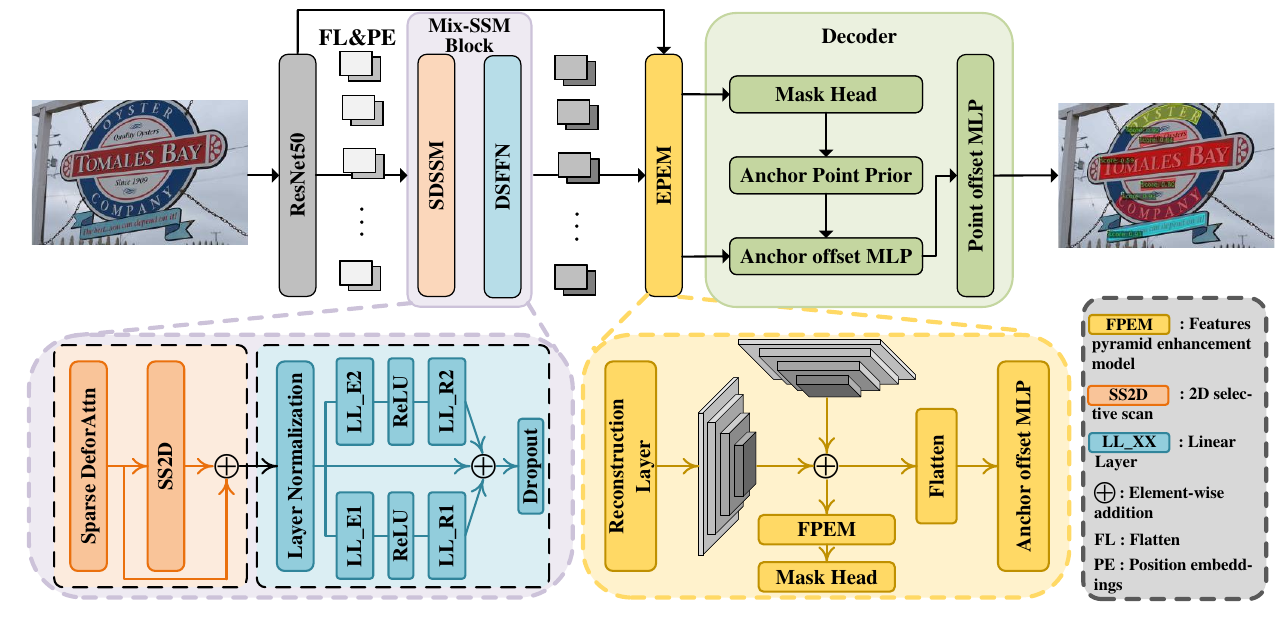}}
\caption{The overall framework of our method.}
\label{fig:res}
%
\end{figure*}

\section{Method}

The overall framework of the proposed model is illustrated in Fig. 2. Our experiments reveal that the straightforward introduction of SS2D results in significant performance degradation. To address this, we carefully designed the framework and components of the model to ensure overall effectiveness. Specifically, ResNet50 is employed as the backbone network to generate feature maps at four different scales, $F_i \in \mathbb{R}^{\frac{W}{2^i} \times \frac{H}{2^i} \times C}$, where $i \in \{2, 3, 4, 5\}$ and $C$ represents the channel dimension. These feature maps $F_i$ are then flattened into embeddings sequences $S \in \mathbb{R}^{L \times C}$, following the approach of DPText-DETR \cite{11}, where $L$ represents the length of the embeddings.

The embedding sequences $S$ are subsequently passed to the encoder’s Mix-SSM blocks to generate contextualized information $\widetilde{S}$. Specifically, the flattened image embeddings $S$ first pass through deformable attention layers to extract global features. This is followed by the application of the Top\_k algorithm, which selects sparsified attention values. The explicitly selected key information is then processed by SS2D to model long-distance dependencies, and high-dimensional information is further modeled through the Dual-Scale Feed Forward Network (DSFFN). Given the importance of multi-scale feature fusion in text segmentation \cite{PSE, 1}, an embedding pyramid enhancement module (EPEM) is designed to fuse multi-scale features within $\widetilde{S}$. The decoder first predicts coarse instance masks and regresses anchor point coordinates, then refines the anchor offsets to accurately determine the polygon control points.

\subsection{Mix-SSM Block}

The Mix-SSM block serves as the fundamental unit of the encoder, comprising two key components: the sparsed deformable state space model (SDSSM) and the dual-scale feed-forward network (DSFFN). Initially, the image embeddings sequences $S$ are passed to the deformable attention layer \cite{10}. To ensure that the model focuses on the most critical information, we employ the $Top\_k$ selection algorithm to sparsify the query-key attention matrix ${\ W}_{Attn}$. Specifically, as illustrated in (1), we use the scatter function to zero out all values in each row of the attention matrix, retaining only the top $k$ values, where $\mathcal{T}_k(\cdot)$ is the $Top\_k$ selection operator.

\begin{small}
\begin{equation}
[\mathcal{T}_k({\ W}_{Attn})]_{i,j}=
\begin{cases}
({\ W}_{Attn})_{i,j}&  ({\ W}_{Attn})_{ij}\in Top\_k(row_j),\\
0&  otherwise
\end{cases}
\end{equation}
\end{small}

We then integrate the SS2D model \cite{13} into the sparse deformable attention layer to enhance long-range feature extraction without significantly increasing computational cost. Details of the SS2D are discussed in Section II. B. SS2D performs selective scanning and feature extraction on the embedding tokens. The number of hidden layers in SS2D is set to 16, following the original parameter settings \cite{14}.

\subsection{Dual-scale Feed Forward Network (DSFFN)}

The DSFFN replaces the traditional feed-forward network \cite{9} to enhance interactions between different hidden layer state representations. As depicted in Fig. 2, we use layer normalization for feature scaling, and the resulting normalized feature is denoted as $L_{in}$. $LL\_E1$ means to expand the channel dimensions from $C$ to $2C$. In contrast to the $LL\_E1$ operation, $LL\_R1$ denotes reduced channel dimensions back to $C$. Similarly, $LL\_E2$ denotes the extension of channel dimensions from $C$ to $4C$, and $LL\_R2$ means reducing them back to $C$. The DSFFN facilitates feature interaction across dual-scale paths as defined in (2), $\oplus$ denotes element-wise addition, and $\sigma[\cdot]$ is the ReLU activation function. Here, $\mathcal{E}_1$ and $\mathcal{E}_2$ correspond to $LL\_E1$ and $LL\_E2$, while $\mathcal{R}_1$ and $\mathcal{R}_2$ correspond to $LL\_R1$ and $LL\_R2$.

\begin{small}
\begin{equation}
L_{out}=L_{in}\oplus \mathcal{R}_1(\sigma[\mathcal{E}_1(L_{in})])\oplus\mathcal{R}_2(\sigma[\mathcal{E}_2(L_{in})])\in\mathbb{R}^{L\times C}
\end{equation}
\end{small}

\subsection{Embedding Pyramid Enhancement Module(EPEM)}

Multi-scale feature fusion has been shown to significantly enhance the robustness of models in detecting arbitrarily shaped text instances. However, this approach is typically applied to two-dimensional image feature maps generated by CNNs. To achieve comparable improvements in one-dimensional image embeddings, we developed an embedding pyramid enhancement module.

In the EPEM model, the encoder’s output $\widetilde{S}$ first generates a reconstructed feature map ${F_i}^r \in \mathbb{R}^{\frac{W}{2^i} \times \frac{H}{2^i} \times C}$, where the height ($\frac{H}{2^i}$) and width ($\frac{W}{2^i}$) correspond to the decomposed embeddings length $L$. The model then fuses ${F_i}^r$ with the backbone feature $F_i$ via element-wise addition, producing a pixel embedding ${F_i}^{'} \in \mathbb{R}^{\frac{W}{2^i} \times \frac{H}{2^i} \times C}$ to mitigate the feature information bottleneck. A feature pyramid enhancement module (FPEM) \cite{1} is used to fuse multi-scale features, retaining only the enhanced feature ${F_3}^{'}$ for input to the mask head. Finally, $F_{i}^{'}$ are flattened again and passed to the decoder as the updated $\widetilde{S}$.

\begin{figure*}[h]
\centering
\centerline{\includegraphics[width=17.5cm]{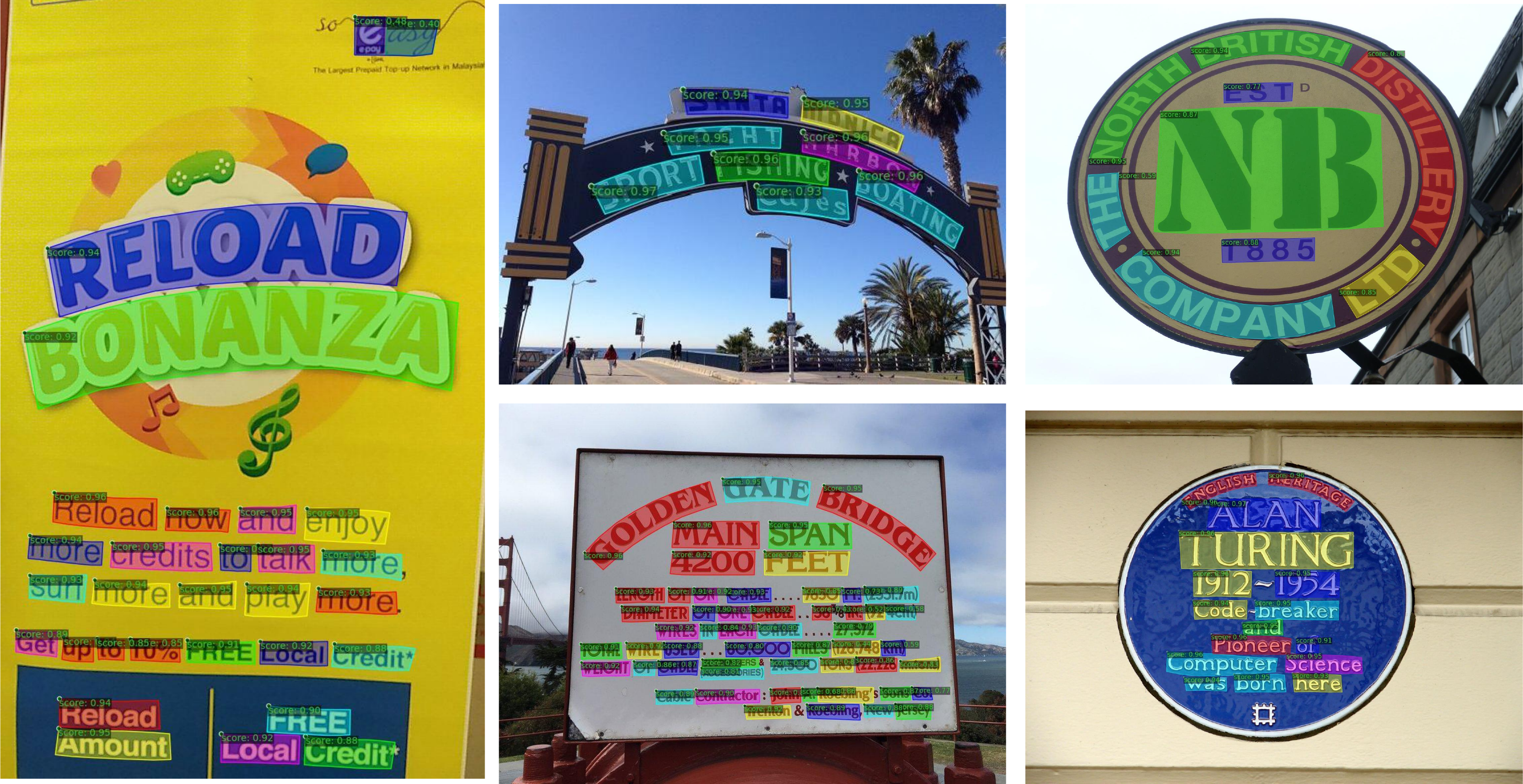}}
\caption{Qualitative results of our method on TotalText.}
\label{fig:res}
%
\end{figure*}

\subsection{Decoder Architecture}

Decoder embeddings $Q$ are initialized with the sequence $\widetilde{S}\in\mathbb{R}^{L\times C}$. To enhance the robustness of control point feature transferability, we introduce learnable embeddings $C=\{c_1,...,c_n\}\in\mathbb{R}^{n\times C}$, where $n$ is the number of polygon control points (16), as shown in (3),

\begin{equation}
Q=\widetilde{S}[\mathcal{T}_k({score}_{cls})]+C\in \mathbb{R}^{K\times n \times C}
\end{equation} 
${score}_{cls}$ is the classification score predicted from $\widetilde{S}$. To obtain the instance-wise feature, $\mathcal{T}_k(\cdot)$ is applied, where $K$ is the number of proposals (100). In the mask head, the mask embedding $Q_m$ is first formulated through weighted summation of $Q$ along the control points dimension. The model then derives the embeddings-wise mask prediction, ${Mask}_e$, using both $Q$ and $Q_m$. The process is as follows:

\begin{small}
\begin{equation}
\begin{cases}
{Mask}_e=Sigmoid[{\rm Conv}_{1\times1}({\rm Conv}_{1\times9}(Q))]\oplus Q_m \in \mathbb{R}^{K\times 1 \times C}\\
{Mask}_i=MLP({Mask}_e)\cdot P(F_{3}^{'})\in \mathbb{R}^{K\times \frac{H}{8} \times \frac{W}{8}}
\end{cases}
\end{equation}
\end{small}

Finally, ${Mask}_e$ performs dot-product with $F_{3}^{'}$ to obtain the instance-wise mask prediction ${Mask}_i$. ${\rm Conv}_{1\times1}(\cdot)$ and ${\rm Conv}_{1\times9}(\cdot)$ are 1D convolutions with 1x1 and 1x9 kernels, respectively, $MLP(\cdot)$ refers to the MLP scaling network, and $P(\cdot)$ denotes linear projection. 

To perform polygon control points regression through mask prediction, we first use the linspace function to generate the anchor grids map $ M \in \mathbb{R}^{\frac{W}{8} \times \frac{H}{8} \times 2}$, matching the resolution of ${Mask}_i$. We then compute the Hadamard product of $M$ with the normalized mask ${Mask}_i$ to obtain the anchor point priors $P=Softmax({Mask}_i) \odot M \in \mathbb{R}^{K\times 2} $. These priors $P$ are then input to the anchor offset $MLP$ as reference points. Finally, the decoder refines the prediction points layer by layer to obtain the regression results.

\subsection{Optimization}

The overall loss function of the framework is as follows:

\begin{align}
L= & \lambda_{cls}FL({score}_{cls}, {score}_{gt})+ \lambda_{seg}Dice({Mask}_i, \\ \nonumber 
 & {Mask}_{gt})+ \lambda_{reg}L1(P^n, P_{gt}^n)
\end{align}

The $\lambda_{cls}$, $\lambda_{seg}$, and $\lambda_{reg}$ are balancing parameters used to adjust the emphasis on different loss components. In our experiments, $\lambda_{cls}$, $\lambda_{seg}$, and $\lambda_{reg}$ are set to 2, 5, and 5, respectively. Focal loss $FL(\cdot)$ is applied to the classification scores ${score}_{cls}$ in the decoder, where ${score}_{gt}$ is the true classification score. Dice loss $Dice(\cdot)$ supervises mask predictions ${Mask}_i$ with ${Mask}_{gt}$ as the ground truth. L1 distance loss $L1(\cdot)$ is used to regress polygon control points with the predicted control points $P^n$ and ground truth $P_{gt}^{n}$.

\section{Experiment}

\subsection{Datasets}

\textbf{SynthText \cite{17}} SynthText is a large-scale, synthesized scene text dataset comprising 149,059 images. The synthesis process involves overlaying text with varying sizes, orientations, and curvatures onto diverse natural scene images as backgrounds. SynthText is used as the dataset for the pre-training stage.

\textbf{MLT17 \cite{18}} MLT17 is a large-scale multilingual scene text dataset comprising 18,000 images. It includes text instances in nine languages, characterized by arbitrary shapes and orientations. MLT17 is utilized as a pre-training dataset to enhance the model's robustness across different languages.

\textbf{TotalText \cite{19}} TotalText consists of 1,500 images of text in natural scenes, including both horizontal and curved text. It is used for model fine-tuning in the experiment.

\textbf{CTW1500 \cite{20}} CTW1500 is a dataset specialized in curved text, commonly used to evaluate a model's robustness to such text. It consists of 1,000 training images and 500 test images.

\textbf{ICDAR19-ArT\cite{21}} ICDAR19-ArT is an arbitrarily shaped scene text dataset, an extension of Total-Text and CTW1500, comprising 5,603 training images and 4,563 test images. The dataset primarily contains multilingual text instances with arbitrary shapes.

\subsection{Implementation details}

The model was pre-trained on SynthText and MLT17 for 200k epochs with a learning rate of 1e-4. It was then fine-tuned on TotalText, CTW1500, and ICDAR19-ArT for 20k epochs with a learning rate of 5e-5. The training strategy and deformable attention layer implementation details follow \cite{11}. Experiments were conducted on a single GTX3090 GPU. Model performance was evaluated using precision (P), recall (R), and F-measure (F). Additionally, we computed the model's parameters and frames per second (FPS) to assess its efficiency.

\begin{table}[h] 
\centering
\caption{Experiment results on CTW1500 and TotalText.}
\resizebox{\linewidth}{!}{

\begin{tabular}{lcccccccccc}
\toprule%
Method & Paper & \multicolumn{3}{c}{CTW1500} & \multicolumn{3}{c}{TotalText}  \\\cmidrule{3-5}\cmidrule{6-8}%
& & R & P & F & R & P & F  \\
\midrule
\multicolumn{8}{c}{CNN-based methods} \\\cmidrule{1-8}
TextSnake\cite{6}  & ECCV’18 & 85.3 & 67.9 & 75.6 & 74.5 & 82.7 & 78.4  \\
CRAFT\cite{22}  & CVPR’19 & 81.1 & 86.0 & 83.5 & 79.9 & 87.6 & 83.6  \\
PAN++\cite{2}  & ICCV’19 & 81.2 & 86.4 & 83.7 & 81.0 & 89.3 & 85.0  \\
ABCNetV2\cite{4}  & TPAMI’21 & 83.8 & 85.6 & 84.7 & 84.1 & 90.2 & 87.0  \\
DBNet++\cite{3}  & TPAMI’22 & 83.2 & 88.9 & 86.0 & 82.8 & 87.9 & 85.3  \\
TCM-DB\cite{tcm}  & CVPR'23 & - & - & 85.1 & - & - & 85.9  \\
EK-Net\cite{23}  & ICASSP'24 & 83.7 & 87.9 & 85.8 & - & - & -  \\
ODM\cite{24}  & CVPR'24 & 85.4 & 85.9 & 85.6 & 83.4 & 88.6 & 85.9  \\
LR-DiffText\cite{LR}  & IJCNN'24 & 81.4 & 88.5 & 84.8 & 78.2 & 80.1 & 79.1  \\
\midrule
\multicolumn{8}{c}{Transformer-based methods} \\\cmidrule{1-8}
I3CL\cite{26}  & IJCV’22 & 84.5 & 87.4 & 85.9 & 83.7 & 89.2 & 86.3  \\
TESTR\cite{testr}  & CVPR’22 & 83.1 & 89.7 & 86.3 & 83.7 & 92.8 & 88.0  \\
TextBPN++\cite{25}  & TMM’23 & 83.8 & 87.3 & 85.5 & 85.3 & 91.8 & 88.5  \\
DPText\cite{11}  & AAAI’23 & 86.2 & \textbf{91.7} & 88.8 & 86.3 & 91.3 & 88.7  \\
DeepSolo\cite{deep}  & CVPR’23 & - & - & - & 84.6 & \textbf{93.2} & 88.7  \\
DTTR\cite{9}  & ICASSP’23 & 84.1 & 85.5 & 84.8 & 73.4 & 86.5 & 79.4  \\
\hline\hline

Ours  & - & \textbf{88.5} & 91.0 & \textbf{89.7} & \textbf{88.8} & 89.5 & \textbf{89.2}  \\
\hline

\end{tabular}
}
\end{table}

\begin{figure*}[h]
\centering
\centerline{\includegraphics[width=17.5cm]{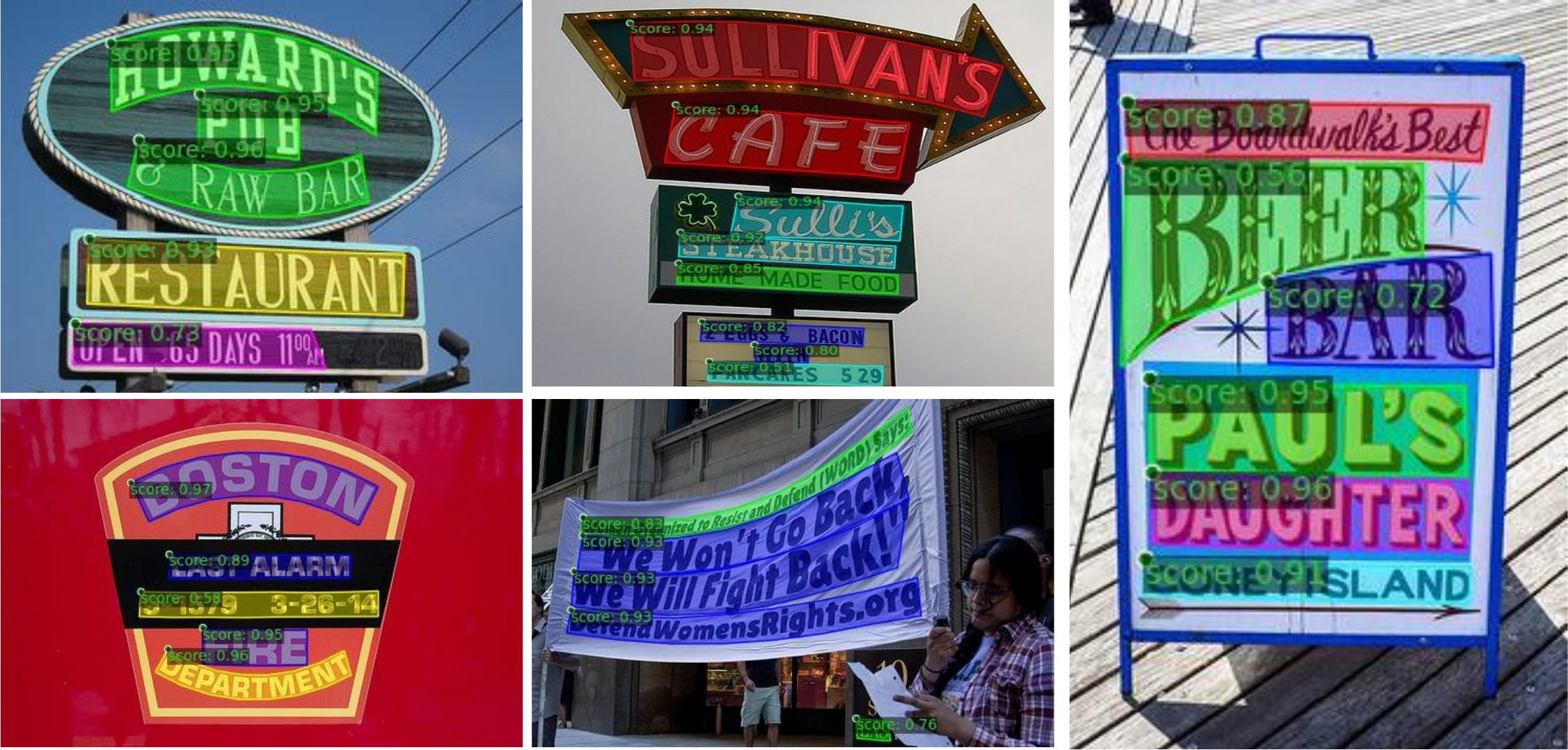}}
\caption{Qualitative results of our method on CTW1500.}
\label{fig:res}
\end{figure*}

\subsection{Text Detection Results on TotalText and CTW1500}

We evaluate our method with current state-of-the-art (SOTA) methods on the CTW1500 and TotalText datasets. Leveraging our well-designed modeling framework and key components, including the Mix-SSM block, EPEM, and DSFFN, our model consistently achieves the best performance across various benchmarks, demonstrating its ability to model features more effective.

Quantitative results are reported in Table I, and qualitative results are shown in Fig. 3 and Fig. 4. Our method achieves SOTA performance on both benchmarks. Comparing our approach with existing CNN-based and Transformer-based methods demonstrates that integrating state space models significantly enhances text detection performance.

Compared to CNN-based methods, our approach surpasses the best-reported method, ABCNetV2, by 2.2\% in F-measure on the TotalText dataset and outperforms DBNet++ by 3.7\% on CTW1500. When compared to existing Transformer-based methods, our method achieves either superior or comparable results. Notably, our approach outperforms the previous study DPText, which uses the deformable attention encoder. Specifically, our method achieves a recall improvement of +2.3\% on CTW1500 and +2.5\% on TotalText and an F-measure improvement of +0.9\% on CTW1500 and +0.5\% on TotalText.

\begin{table}[h] 
\centering
\caption{Quantitative results on ICDAR19 ArT.}
\resizebox{\linewidth}{!}{
\begin{tabular}{lccccc}
\toprule%
Method & Paper & Backbone & \multicolumn{3}{c}{ICDAR19 ArT}  \\\cmidrule{4-6}%
& & & R & P & F  \\
\midrule
PAN++\cite{2}  & ICCV’19 & Res18 & \textbf{79.4} & 61.1 & 69.1\\
CRAFT\cite{22}  & CVPR’19 & VGG16 & 68.9 & 77.2 & 72.9\\
TextBPN++\cite{23} & TMM’23 & Res50 & 71.1 & 81.1 & 75.8 \\
I3CL\cite{26} & IJCV’22 & Res50 & 71.3 & 82.7 & 76.6  \\
DPText\cite{11} & AAAI’23 & Res50 & 73.7 & 83.0 & 78.1  \\
\midrule
Ours  & - & Res50 & 71.9 & \textbf{86.1} & \textbf{78.5}\\
\hline

\end{tabular}
}
\end{table}

\subsection{Text Detection Results on ICDAR19 ArT}

The proposed method was also evaluated against state-of-the-art methods on the ICDAR19 ArT dataset, which contains multilingual and arbitrarily shaped scene text. As shown in Table II, our method demonstrates a superior balance between recall and precision. Additionally, it achieves the highest precision (P) and F-measure on the ICDAR19 ArT dataset, with values of 86.1\% and 78.5\%, respectively. Qualitative results are presented in Fig. 6, highlighting the robustness of our method to different languages, arbitrary shapes, and dense text.

\subsection{Ablation studies}
\label{ssec:subhead}

\begin{table}[h]
\centering
\caption{Results of ablation experiments with SS2D.}
\resizebox{\linewidth}{!}{

\begin{tabular}{cccccccc}
\toprule
\#Attn Layer & \#SS2D & P & R & F & Params(M)\\
\midrule
6 & \checkmark & 90.98 & 88.54  & 89.74 & 5444\\
6  & -  & 85.73 & 88.95 & 87.31 &5103\\
-  & \checkmark  & 85.46 & 83.52 & 84.48 &4758\\
12  & -  & 87.32 & 88.90 & 88.11 &5558\\
18  & -  & 89.42 & 89.26 & 89.34 &6012\\
\bottomrule
\end{tabular}
}
\end{table}

\subsubsection{Effectiveness of SS2D}

The encoder of the baseline model consists of six deformable attention layers. To investigate how Mamba can enhance the encoder's capacity to extract relevant information from long sequences, we initially adopted a straightforward approach by directly replacing the attention layers with SS2D. However, this approach led to a significant performance degradation, with the F-measure decreasing by 2.83\%. Consequently, we decided to retain the original attention layers to maintain the integrity of the DETR framework, while employing a more refined strategy to incorporate SS2D effectively.

As shown in Table III, our approach demonstrates that SS2D further enhances model performance, resulting in an improvement of 2.43\% in F-measure. Additionally, the Mix-SSM model incorporating SS2D achieves an optimal F-measure of 89.74, albeit with a slight increase in parameters. Specifically, when the number of attention layers is increased to 18, the F-measure reaches 89.34, but the number of parameters rises by 909M compared to the baseline. In contrast, the new structure with SS2D delivers a higher F-measure, with a parameter increase that is only 37.5\% of that seen with the 18-attention-layer configuration.

By incorporating SS2D with a selection mechanism, the model’s ability to capture long-range feature dependencies is significantly enhanced, thereby improving its detection and perception of text instances in scenes with extreme scales or arbitrary shapes. To further enhance model interpretability, we present visualizations of the detection and attention results in Fig. 5, which demonstrate that SS2D improves the model's robustness to complex-scale text instances.

\begin{figure}[h]
\centering
\centerline{\includegraphics[width=7.7cm]{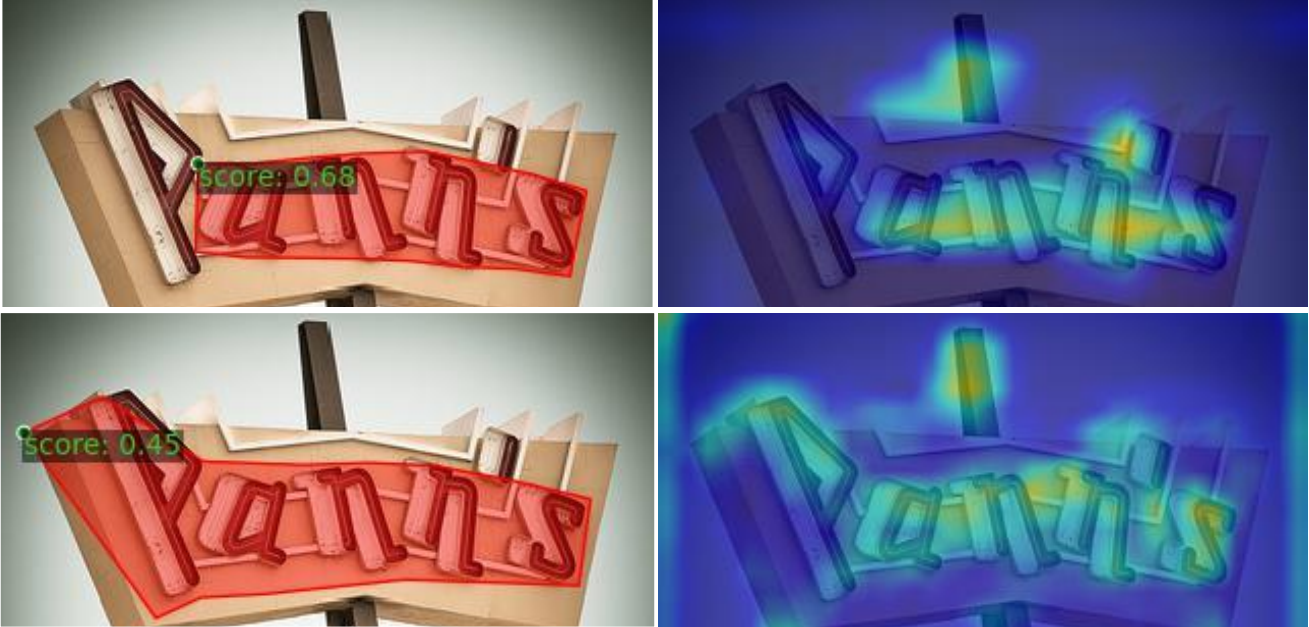}}
\caption{Visualization of detection and attention results.}
\label{fig:res}
\end{figure}

\begin{table}[h]

\centering
\caption{Detection results under different settings.}

\begin{tabular}{ccccccc}
\toprule
EPEM & DSFFM & FPS & P & R & F \\
\midrule
- & - & 7.2 & 87.38 & 90.44 & 88.88\\
\checkmark  & - & 7.0 &  90.98 & 87.46 & 89.19\\
-  & \checkmark & 7.1 & 90.40 & 87.95 & 89.16\\
\checkmark  & \checkmark & 6.9 & 90.98 & 88.54 & 89.74\\
\bottomrule
\end{tabular}

\end{table}

\begin{figure*}[h]
\centering
\centerline{\includegraphics[width=17.5cm]{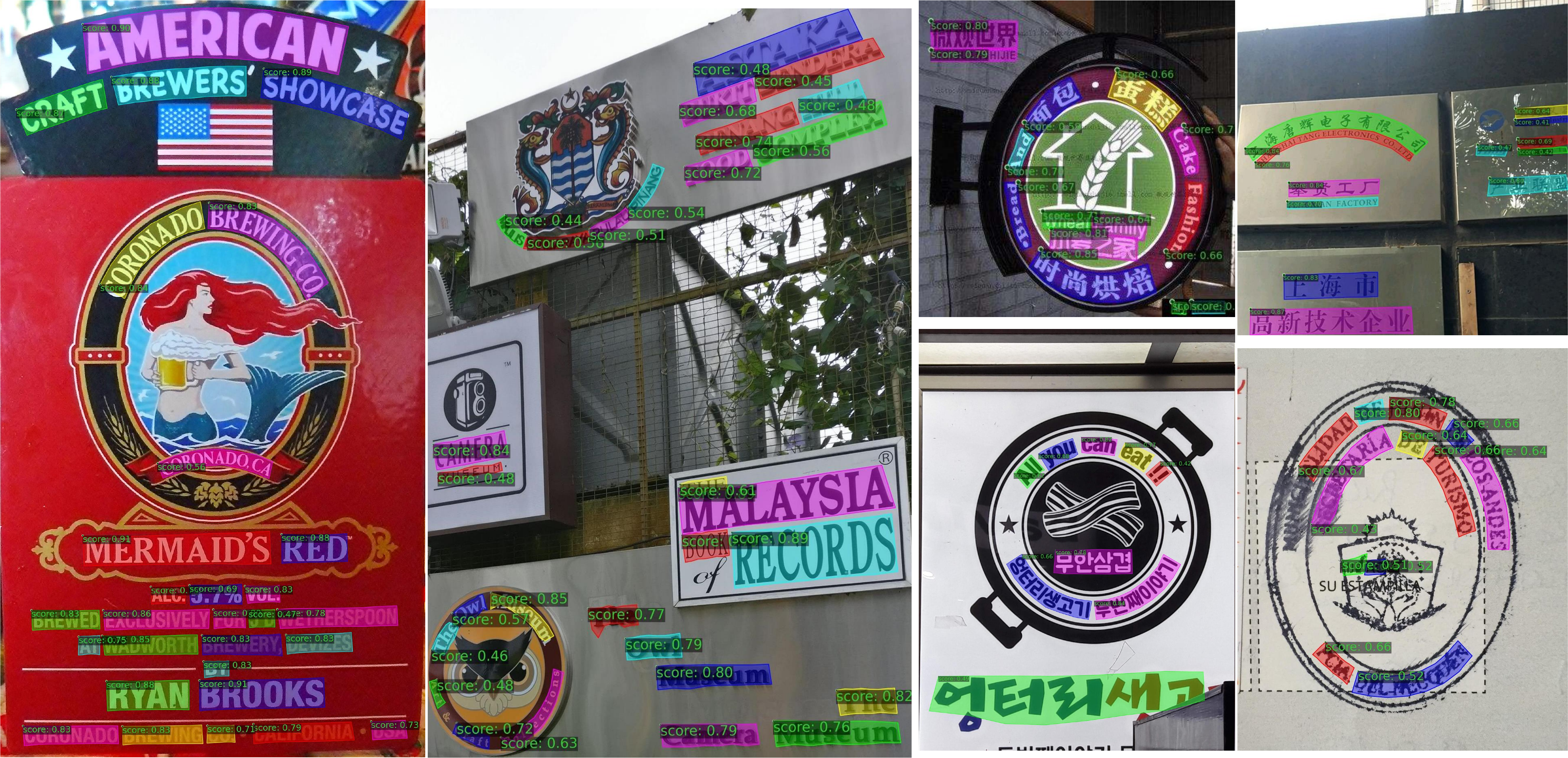}}
\caption{Qualitative results of our method on ICDAR19-ArT.}
\label{fig:res}
\end{figure*}

\subsubsection{Effectiveness of EPEM}

This paper introduces the Embedding Pyramid Enhancement Module (EPEM) to address the token sequence feature information bottleneck through fine-grained multi-scale feature fusion. As shown in Table IV, EPEM reduces false positive predictions by fusing multi-scale features, resulting in a 3.6\% improvement in precision. Furthermore, the F-measure improves by 0.31\%. Additionally, testing of the model's runtime speed reveals that the EPEM module is lightweight, contributing only a 0.2 FPS increase.

\subsubsection{Effectiveness of DSFFN}

To verify the effectiveness of DSFFN, the baseline used a vanilla single-path feed-forward network as the control setting for the ablation experiments. The results in Table IV show that introducing DSFFN leads to at least a 0.28\% improvement in F-measure. Moreover, the combined use of DSFFN and EPEM synergistically enhances the F-measure by 0.86\%. Additionally, DSFFN is designed to be lightweight, resulting in an FPS drop of only 0.1.

\section{Conclusion}

In this paper, we analyze the limitations of Transformer-based scene text detection methods, particularly their insufficient capability to model long-range dependencies. To address this, we propose TextMamba, which incorporates the selection mechanism of Mamba to enhance the model’s ability to capture long-range information. A Mix-SSM block is designed to effectively integrate the selection mechanism with deformable attention layers. To mitigate the interference of irrelevant information in long-range modeling, we employ the Top\_k algorithm to sparsify attention values and explicitly select key information. Additionally, we introduce a dual-scale Feed Forward Network (DSFFN) to improve information interaction and an embedding pyramid enhancement module to enable token-wise multi-scale feature fusion. Experimental results demonstrate that the proposed method outperforms the previous state-of-the-art approaches on three benchmarks. While our approach shows promising performance, further exploration of more lightweight frameworks remains necessary, which we have identified as a direction for future research.

\section{Acknowledgement}

This work is supported by the Open Project of the State Key Laboratory of Multimodal Artificial Intelligence Systems (MAIS2024101), the Unveiling and Leading Projects of Xiamen (No. 3502Z20241011), and the Major Science and Technology Plan Project on the Future Industry Fields of Xiamen City (No. 3502Z20241027).

\end{document}